\documentclass[11pt,a4paper]{article}
\usepackage[hyperref]{emnlp2020}
\usepackage{times}
\usepackage{latexsym}
\usepackage{hyperref}
\usepackage{graphicx}
\usepackage{tikz}
\usepackage{multirow}
\usepackage{array}
\usepackage{makecell}
\usepackage{graphicx}
\usepackage{wrapfig}
\usepackage{amsmath}
\usepackage{amsthm}
\usepackage{enumitem}
\usepackage{colortbl}
\definecolor{mycolor}{rgb}{197, 225, 184}

\usepackage{microtype}

\aclfinalcopy 

\title{Language Models as Few-Shot Learner for \\ Task-Oriented Dialogue Systems}

\author{Andrea Madotto, Zihan Liu, Zhaojiang Lin, Pascale Fung \\
  Center for Artificial Intelligence Research (CAiRE)\\
  The Hong Kong University of Science and Technology\\
  \texttt{amadotto@connect.ust.hk} }

\date{}

\begin{document}
\maketitle
\begin{abstract}
Task-oriented dialogue systems use four connected modules, namely, Natural Language Understanding (NLU), a Dialogue State Tracking (DST), Dialogue Policy (DP) and Natural Language Generation (NLG). A research challenge is to learn each module with the least amount of samples (i.e., few-shots) given the high cost related to the data collection. The most common and effective technique to solve this problem is transfer learning, where large language models, either pre-trained on text or task-specific data, are fine-tuned on the few samples. These methods require fine-tuning steps and a set of parameters for each task. Differently, language models, such as GPT-2~\cite{radford2019language} and GPT-3~\cite{brown2020language}, allow few-shot learning by priming the model with few examples. In this paper, we evaluate the priming few-shot ability of language models in the NLU, DST, DP and NLG tasks. Importantly, we highlight the current limitations of this approach, and we discuss the possible implication to future work. 
\end{abstract}

\section*{Acknowledgments}
I would like to thanks \href{https://jasonwu0731.github.io/}{Jason Wu} for providing an easy to use code in ToD-BERT and for clarification about the code and tasks, \href{https://scholar.google.com/citations?user=u1CNjgwAAAAJ&hl=zh-CN}{Baolin Peng} for the easy to use repository FewShotNLG and for providing help with the scorer, and \href{https://dathath.github.io/}{Sumanth Dathathri} for the discussion and insight about the limitation of the LM priming few-shots. 

\section{Introduction}
Modularized task-oriented dialogues systems are the core of the current smart speaker generation (e.g., \href{https://en.wikipedia.org/wiki/Amazon_Alexa}{Alexa}, \href{(https://en.wikipedia.org/wiki/Siri}{Siri} etc.). The main modules of such systems are Natural Language Understanding (NLU), Dialogue State Tracking (DST), Dialogue Policy (DP) and Natural Language Generation (NLG), each of which is trained separately using supervised and/or reinforcement learning. Thus a data collection process is required, which for some of the tasks can be laborious and expensive. For example, dialogue policy annotation has to be done by an expert, better by a professional linguist. Therefore, having a model that requires only few samples to actually perform well in the tasks is essential. 

\begin{figure}[t]
    \centering
    \includegraphics[width=0.75\linewidth]{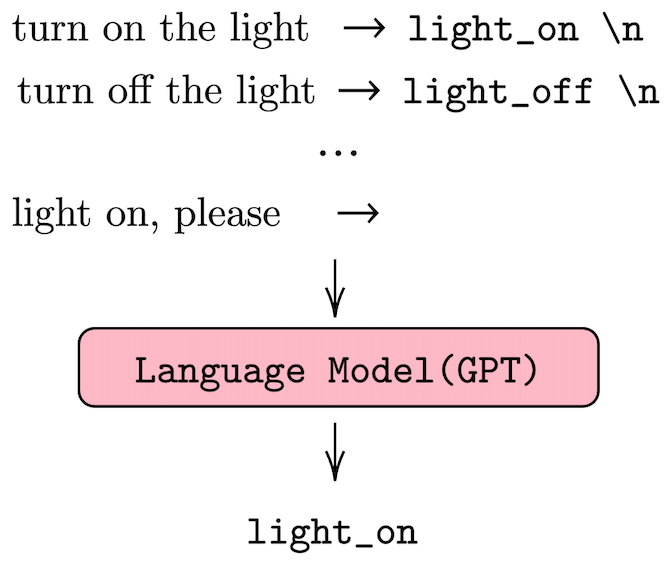}
    \caption{Language model priming for few-shot intent recognition. Image inspired by OpenAI GPT-3~\cite{brown2020language}. Few examples are provided along with the sample to be predicted as the prefix to the language model.}
    \label{fig:main img}
\end{figure}

The most successful approach in few-shot learning for task-oriented dialogue systems is notably transfer learning, where a large model is firstly pre-trained on a large corpus to be then fine-tuned on specific tasks. For task-oriented dialogue systems, ~\citet{wu2020tod} proposed \href{https://github.com/jasonwu0731/ToD-BERT}{TOD-BERT} a large pre-trained model which can achieve better performance than BERT~\cite{devlin2019bert} in few-shots NLU, DST and DP. \citet{liu2020coach} proposed a two-step classification for few-shot slot-filling, a key task for the NLU module. Similarly, \citet{peng2020few} introduced a benchmark for few-shot NLG and a pre-trained language model (\href{https://github.com/pengbaolin/SC-GPT}{SC-GPT}) specialized for the task. Further, a template rewriting schema based on
T5~\cite{raffel2019exploring} was developed by \citet{kale2020few} for few-shot NLG in two well-known datasets. \citet{peng2020soloist} proposed a pre-trained language model (LM) for end-to-end pipe-lined task-oriented dialogue systems. In their experiments, they showed promising few-shot learning performance in MWoZ~\cite{budzianowski2018multiwoz}. Finally, several meta-learning approaches have been proposed for DP~\cite{xu2020meta}, NLG/ACT~\cite{mi2019meta}, pipelined end-to-end models~\cite{qian2019domain} and personalized dialogue systems~\cite{madotto2019personalizing}.

For performing few-shot learning, existing methods require a set of task-specific parameters since the model is fine-tuned with few samples. Differently, in this paper, we perform few-shot learning by priming LMs with few-examples~\cite{radford2019language,brown2020language}. In this setting, no parameters are updated, thus allowing a single model to perform multiple tasks at the same time. In this paper, we evaluate the few-shot ability of LM priming on the four task-oriented tasks previously mentioned (i.e., NLU, DST, DP, and NLG). Currently, GPT-3~\cite{brown2020language} is not available to the public; thus we experiment on different sizes GPT-2~\cite{radford2019language} models such as SMALL (117M), LARGE (762M), and XL (1.54B). All the experiments are run on a single NVIDIA 1080Ti GPU.

\section{Basic Notation and Tasks}
Let us define dialogue as the alternation of utterances between two speakers denoted by $U$ and $S$ respectively. An utterance is a sequence of words $X = {x_1, \cdots , x_n}$ and the concatenation of $t$ utterances denotes a dialogue with $\frac{t}{2}$ turns. In this paper, we focus on the four task-oriented dialogue system tasks, and we briefly introduce the input-output of each task.
\paragraph{NLU} This task aims to extract slot-value pairs (\texttt{SLOT-FILLING}) and the intent (\texttt{INTENT}) from a user utterance $S$. In the literature, the most common approach for NLU is to learn a BIO tagger for the slot-value pairs, and to learn a multi-class classifier for the intent. \texttt{SLOT-FILLING} gets as input a user utterance $X$ and produces a dictionary $M=\{s_1=v_1,\cdots, s_n=v_n\}$, where $s_i$ is a slot and $v_i$ is the possible value. Note that $v_i$ can also be \texttt{None} since some slots may not be mentioned in the utterance. The \texttt{INTENT} task gets a user utterance $X$ and classifies it into an intent class denoted by $Y \in \{y_1,\cdots,y_n\}$. Sometimes, the intent-classification is mixed with the domain classification. 
\paragraph{DST} This task extracts slot-value pairs for a given dialogue, which can be considered as a dialogue-level of the NLU. Given a dialogue with $t$ turns as a sequence of utterance $D=\{X^1_U, X^1_S, \cdots , X^t_U\}$ a DST model predicts a dictionary $M_t=\{s_1=v_1, \cdots, s_n=v_n\}$ as in the NLU. Note that most of the existing DST models use the previously generated $M_{t-1}$ and just update the slots required using an NLU tagger. 
\paragraph{ACT} This task predicts the next speech-act (e.g., INFORM, REQUEST etc.) given the current dialogue state, in the form of a dialogue or dictionary of slot-value pairs. This is usually stated as a reinforcement learning task in both online and offline settings. In this paper, we simplify the tasks, and instead of learning a dialogue policy, we perform dialogue act classification. This a multi-label classification task, since more than one speech-act can be used in an utterance. This task gets as input a user utterance $X$ and classifies it in to a set of possible speech-acts in $I \in \{ I_1, \cdots, I_n \}$.
\paragraph{NLG} This task maps a dialogue-act, which is made of a speech-act plus a dictionary of slot-value pairs, into natural language. The model gets as input a speech-act concatenated with a slot-value dictionary overall denoted as $I (s_1=v_1, \cdots, s_n=v_n)$ and it generates as output an utterance $X$.

In the few-shot setting, a small number of input-output pairs is provided to the model, expecting a high degree of generalization. 

\section{Priming the LM for few-shot learning}
Differently from fine-tuning, few-shot learning with LMs requires designing prefixes to perform few-shot learning. In our four tasks, we use three categories of prefixes: \textit{binary}, \textit{value-based} and \textit{generative}. In the following notation, we use $X$ to represent a generic input and $X_i$ for the $i$-th shot samples, thus implying that the prefix remains fixed during the inference and $X$ can become any input. These prefixes are provided to the LM and the generate tokens become the actual prediction, Figure~\ref{fig:main img} show an example of intent recognition. 
\paragraph{Binary} prefixes are used for classification (namely for intent-classification and speech-act detection). We treat every classification as binary, even multi-class. To perform the few-shot priming, we use the following prefix:
\begin{equation}
    X_1 \rightarrow \texttt{True} \ X^*_1 \rightarrow \texttt{False} \ \cdots \ X \rightarrow 
\end{equation}
where $X_i$ one of the few-shot samples and $X^*_i$ is from other classes or from the false class if it exists. To predict $n$ classes, a set of $n$ prefixes is used and thus $n$ forwards is required. 

\paragraph{Value-based} prefixes are used to assign the value of a certain slot given an utterance, or \texttt{None} if no value is provided. We define a prefix for each slot, similar to TRADE \cite{wu2019transferable}, which requires forwarding the model $n$ times for decoding $n$-slots. To perform the few-shot priming of \textit{one} slot $s$, we use the following prefix:
\begin{equation}
    X_1 \rightarrow s=v_1 \ X^*_1 \rightarrow s=\texttt{None} \ \cdots \ X \rightarrow s= 
\end{equation}
where $v_1$ is the assigned value from the few-shot training. This process is repeated for each slot to generate the dictionary $M$.

\paragraph{Generative} prefixes are used to instruct the model to generate natural language given source information (e.g., NLG). The prefix is the following: 
\begin{equation}
    X_1 \rightarrow  Y_1  \ \cdots \ X_k \rightarrow  Y_k \  X \rightarrow 
\end{equation}
where $X_i$ and $Y_i$ are generic sequences of words. 

\begin{figure}[t]
    \begin{minipage}[t]{0.4\textwidth}
      \vspace{0pt}\raggedright
       \centering
        \resizebox{!}{0.11\linewidth}{
        \begin{tabular}{l}
        \rowcolor{mycolor} 
        \begin{tabular}[c]{@{}l@{}}\texttt{turn on the light $\rightarrow$ name=None} \\\texttt{add to playlist kojak $\rightarrow$ name=kojak}\\\texttt{add tune to my hype playlist $\rightarrow$ name=}\end{tabular} \\
        \end{tabular}
        }
    \end{minipage}
    \hfill
    \begin{minipage}[t]{0.48\textwidth}
          \vspace{3pt}
        \includegraphics[width=\linewidth]{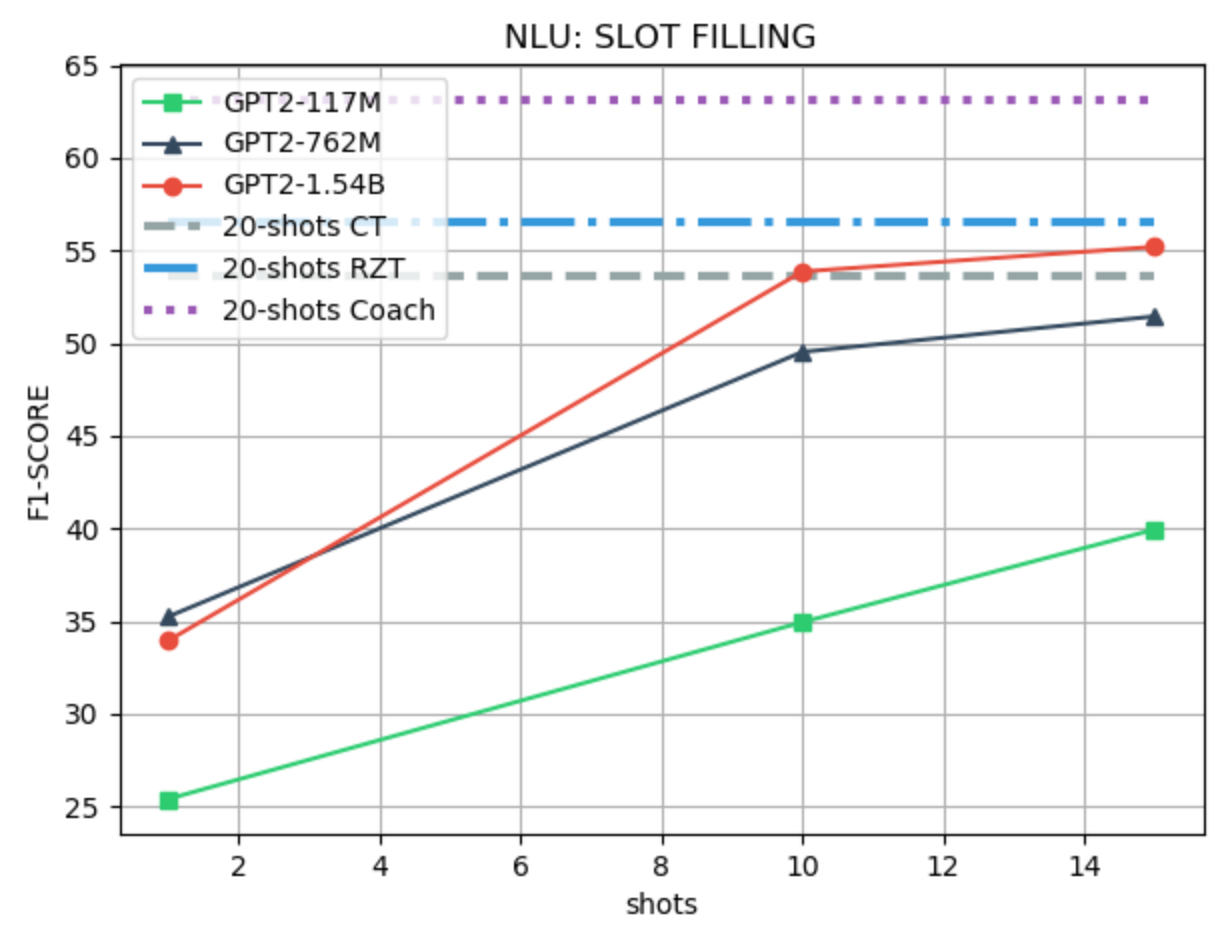}
        \caption{Example of 1-shot LM priming for the \texttt{SLOT-FILLING} task and results in the task. CT, RZT, Coach are from \cite{liu2020coach} and they use 20-shots.} 
       \label{fig:slot-filling}
    \end{minipage}
\end{figure} 	

\begin{figure}[t]
    \begin{minipage}[t]{0.4\textwidth}
      \vspace{0pt}\raggedright
       \centering
        \resizebox{!}{0.09\linewidth}{
        \begin{tabular}{l}
        \rowcolor{mycolor} 
        \begin{tabular}[c]{@{}l@{}}\texttt{yes, your booking is successful $\rightarrow$ booked=True} \\\texttt{what type of food? $\rightarrow$ booked=False}\\\texttt{i do not seem to be finding anything $\rightarrow$ booked=}\end{tabular} \\
        \end{tabular}
        }
    \end{minipage}
    \hfill
    \begin{minipage}[t]{0.48\textwidth}
          \vspace{3pt}
        \includegraphics[width=\linewidth]{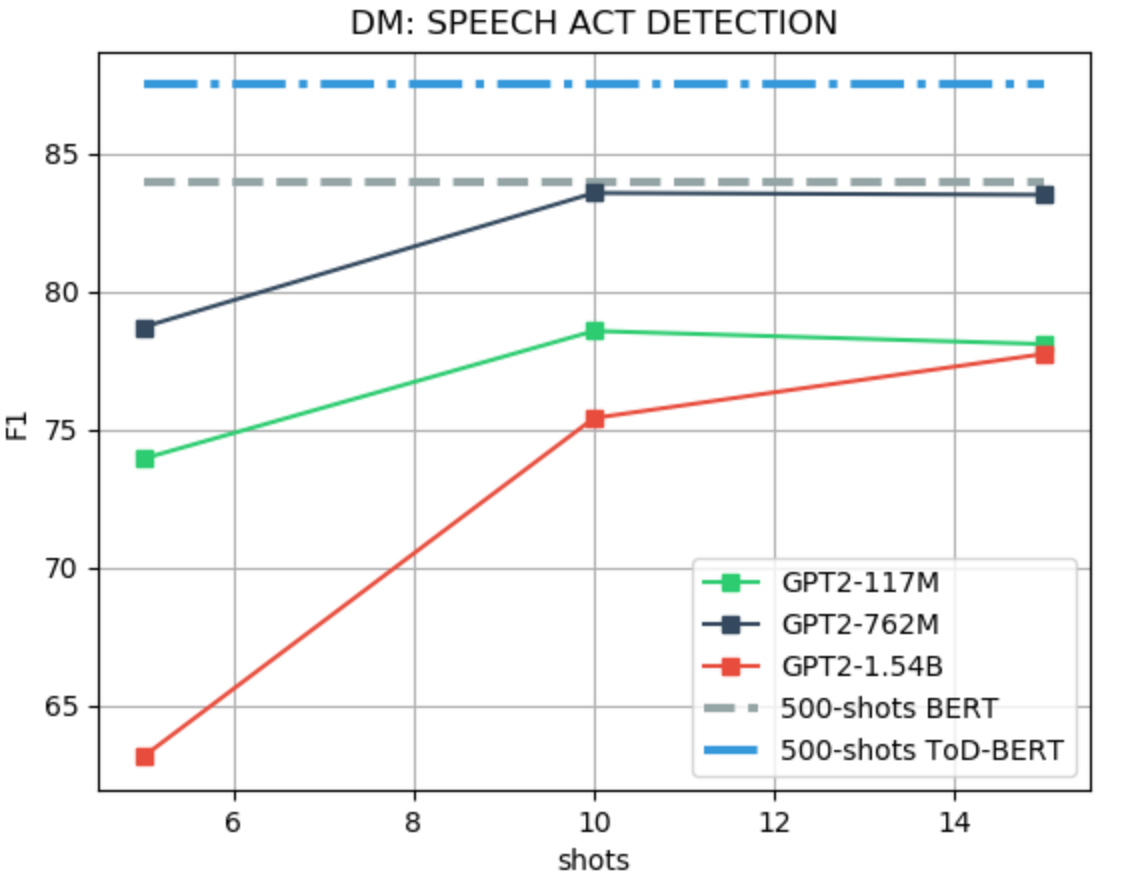}
        \caption{Example of 1-shot LM priming for the \texttt{ACT} task and results in the task.  BERT and ToD-BERT are from \cite{wu2020tod} and they use 500-shots.} 
       \label{fig:ACT}
    \end{minipage}
\end{figure}

\begin{figure}[t]
    \begin{minipage}[t]{0.4\textwidth}
      \vspace{0pt}\raggedright
       \centering
        \resizebox{!}{0.095\linewidth}{
        \begin{tabular}{l}
        \rowcolor{mycolor} 
        \begin{tabular}[c]{@{}l@{}}\texttt{inform(name=hilton...)$\rightarrow$the hilton ...} \\\texttt{inform(name=ocean park...)$\rightarrow$the phone number}\\\texttt{inform(name='super 8...)$\rightarrow$}\end{tabular} \\
        \end{tabular}
        }
    \end{minipage}
    \hfill
    \begin{minipage}[t]{0.48\textwidth}
          \vspace{3pt}
          \centering
        \includegraphics[width=0.9\linewidth]{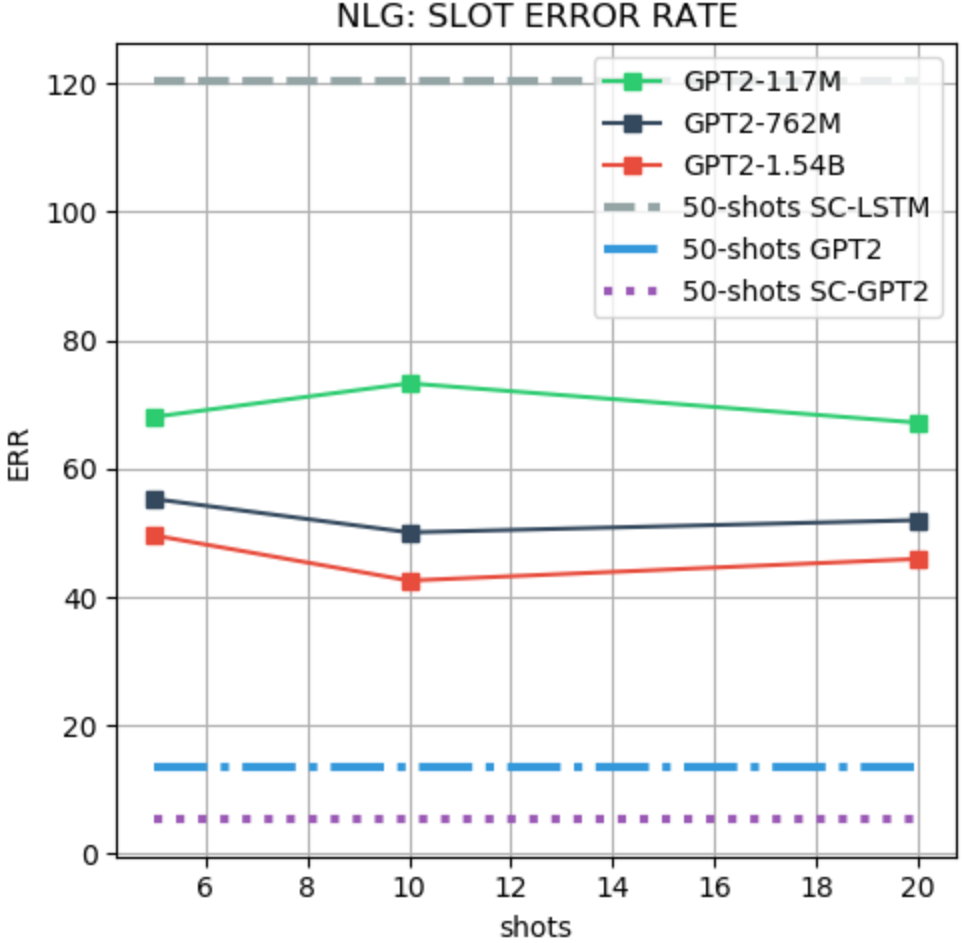}
        \caption{Example of 1-shot LM priming for the \texttt{NLG} task and results in the task. SC-LSTM, GPT-2, and SC-GPT-2 are from \citet{peng2020few}.} 
       \label{fig:NLG}
    \end{minipage}
\end{figure} 	

\section{Experiments and Results}
We use different prefix styles depending on the task and we compare the results of LM few-shot priming with those of the existing finetuning-base models. In all the experiments, we use different number of shots since different tasks may fit more or fewer samples in the 1024 max input size of GPT-2.

\paragraph{NLU} We use the SNIPS~\cite{coucke2018snips} dataset for evaluating the \texttt{SLOT-FILLING} and \texttt{INTENT} recognition tasks. For the \texttt{SLOT-FILLING} task, we follow the few-shot setting of \citet{liu2020coach}, and we use the official CoNLL F1 scorer as the evaluation metric. For the \texttt{INTENT} classification, we fine-tune RoBERTa~\cite{liu2019roberta} with 10 samples and use accuracy as the evaluation metric. We use a \textit{value-based} LM prefix for the \texttt{SLOT-FILLING} task with a maximum of 15 shots, and \textit{binary} LM prefix for the  \texttt{INTENT} classification task with a maximum of 10 shots. An example of a prefix for the \texttt{SLOT-FILLING} task and the few-shot performance evaluation are shown in Figure~\ref{fig:slot-filling}. Table~\ref{tab:slot-filling} and ~\ref{tab:intent} and Figure~\ref{figAPP:NLU} show more detailed results. 

\paragraph{DST} We use the MultiWoZ~\cite{budzianowski2018multiwoz,eric2019multiwoz,zang2020multiwoz} dataset for evaluating the \texttt{DST} task. Differently from other works, we use the last user utterance only as input to the model, and we update the predicted-DST through turns. For the few-shot evaluation, we follow the setting of \citet{wu2020tod}, and we report the joint and slot accuracy. As baselines, we use TOD-BERT~\cite{wu2020tod} and BERT~\cite{devlin2019bert} fine-tuned with 10\% of the training data, which is equivalent to 500 examples. We use a \textit{value-based} LM prefix, as for the \texttt{SLOT-FILLING} task, with a maximum of 15 shots due to limited context. Table~\ref{tab:DST} and Figure~\ref{figAPP:DST} show more detailed results. 

\paragraph{ACT} We use the MultiWoZ dataset for evaluating the speech \texttt{ACT} identification task. Differently from other works, only the system utterance is used as input to the model, instead of including the dialogue history and the user utterance as in \citet{wu2020tod}. For the few-shot evaluation, we follow the setting of \citet{wu2020tod}, i.e., F1-score. As baselines, we use TOD-BERT~\cite{wu2020tod} and BERT~\cite{devlin2019bert}, fine-tuned with 10\% of the training data, which is equivalent to 500 examples. We use a \textit{binary} LM prefix, as for the intent classification task, with a maximum of 15 shots due to limited context. An example of a prefix for the \texttt{ACT} tasks and the few-shot performance evaluation is shown in Figure~\ref{fig:ACT}. Table~\ref{tab:act} and Figure~\ref{figAPP:ACT} show more detailed results. 

\paragraph{NLG} We use the FewShotWOZ~\cite{peng2020few} dataset for evaluating the \texttt{NLG} task. For the few-shot evaluation, we follow the setting of ~\citet{peng2020few} and use the BLEU and slot error rate (SLR) as metrics. We use SC-LSTM, GPT-2, and SC-GPT-2~\cite{peng2020few} as baselines, all fine-tuned with 50 examples from the training data. We use a \textit{generative} LM prefix with a maximum of 20 shots due to limited context. An example of prefix for the \texttt{NLG} task and the few-shot performance evaluation is shown in Figure~\ref{fig:NLG}. Table~\ref{tab:NLG_BLEU} and \ref{tab:NLG_SLR}, and Figure~\ref{figAPP:NLG} show more detailed results.

\section{Analysis and Limitation}
From the experimental results, we observe that: 
\begin{itemize}[leftmargin=*]
    \item The larger the model the better the performance in both the \texttt{NLU} and \texttt{NLG} tasks, while, instead, in the \texttt{DST} and \texttt{ACT} tasks, GPT-2 LARGE (762M) performs better than the XL (1.54B) version. This is quite counterintuitive given the results reported for GPT-3. Further investigation is required to understand whether changing the prefix can help to improve the performance of larger models;
    \item In the \texttt{NLU}, \texttt{ACT} and \texttt{NLG}, LM priming few-shot learning shows promising results, achieving similar or better performance than the weakest finetuning-based baseline, which also uses a larger number of shots. On the other hand, in \texttt{DST} the gap with the existing baseline is still large.
\end{itemize}
We also observe two limitations of the LM priming: 
\begin{itemize}[leftmargin=*]
    \item Using \textit{binary} and \textit{value-based} generation requires as many forwards as the number of classes or slots. Although these forward passes are independent, achieving few-shot learning this way is not as effective as directly generating the class or the tag (e.g., \texttt{NLU}). In early experiments, we tried to covert all the tasks into a \textit{generative} format, thus making the model directly generate the sequence of tags or the class label. Unfortunately, the results in the \textit{generative} format were poor, but we are unsure if larger LMs such as GPT-3 can perform better.
    \item The current max-input length of GPT-2 (1024 tokens) greatly limits the number of shots that can be provided to the model. Indeed, in most of the tasks, no more than 15 shots can be provided, thus making it incomparable with existing models that use a larger number of shots. 
\end{itemize}

\section{Conclusion}
In this paper, we demonstrate the potential of LM priming few-shot learning in the most common task-oriented dialogue system tasks (NLU, DST, ACT and NLG). Our experiments show that in most of the tasks larger LMs are better few-shot learners, confirming the hypothesis in~\citet{brown2020language} and, in some cases, they can also achieve similar or better results than the weakest finetuning-based baseline. Finally, we unveil two limitations of the current LM priming few-shot learning the computational cost and the limited word context size. In future work, we plan to benchmark dialogue-specific models (e.g., DialGPT) and LM with longer context size (e.g., Transformer XL~\cite{dai2019transformer}, LongFormer~\cite{beltagy2020longformer}, and BigBird~\cite{zaheer2020big} etc.). We also plan to investigate adversarial triggers~\cite{wallace2019universal} for improving the few-shot ability of LMs, and to benchmark end-to-end dialogue tasks. 

\bibliography{anthology,emnlp2020}
\bibliographystyle{acl_natbib}

\appendix

\section{Appendices}
\label{sec:appendix}

\begin{figure*}[t]
    \begin{minipage}[t]{\textwidth}
      \vspace{0pt}\raggedright
       \centering
       
        \texttt{SLOT-FILLING}
        \resizebox{!}{0.05\textwidth}{
        \begin{tabular}{l}
        \rowcolor{mycolor} 
        \begin{tabular}[c]{@{}l@{}}\texttt{add tune to my hype playlist $\rightarrow$ entity\_name = None} \\\texttt{add to playlist confidence boost here comes $\rightarrow$ entity\_name = here comes}\\\texttt{add the track bg knocc out to the rapcaviar playlist $\rightarrow$ entity\_name =}\end{tabular} \\
        \end{tabular}
        }
        
        \texttt{INTENT}
        \resizebox{!}{0.05\textwidth}{
        \begin{tabular}{l}
        \rowcolor{mycolor} 
        \begin{tabular}[c]{@{}l@{}}\texttt{listen to westbam alumb allergic on google music $\rightarrow$ playmusic = true} \\\texttt{rate this novel 4 points out of 6 $\rightarrow$ playmusic = false}\\\texttt{add sabrina salerno to the grime instrumentals playlist $\rightarrow$ playmusic =}\end{tabular} \\
        \end{tabular}
        }
    \end{minipage}
    \hfill
    \begin{minipage}[t]{\textwidth}
          \vspace{3pt}
        \includegraphics[width=\linewidth]{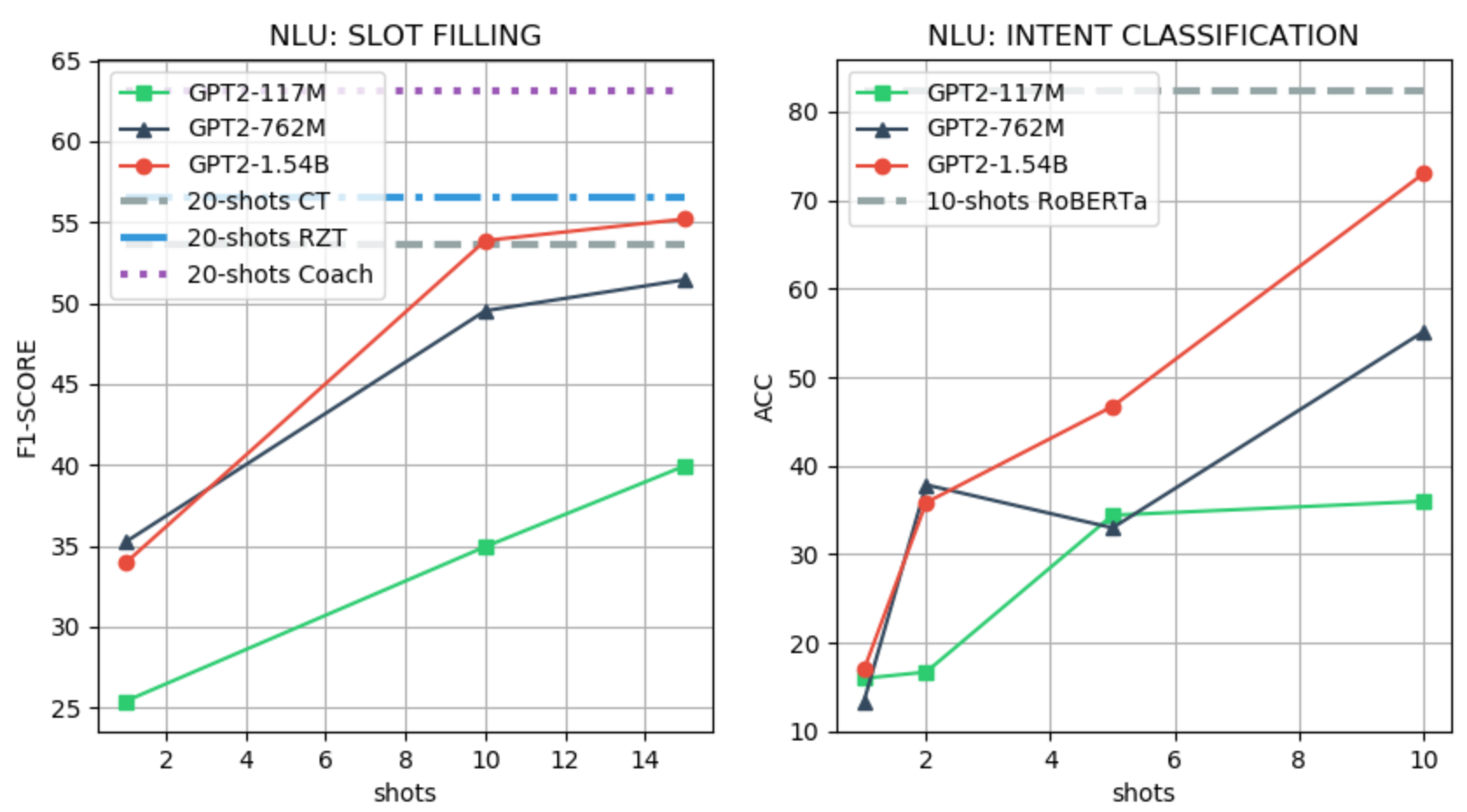}
        \caption{Example of 1-shot LM priming for the \texttt{SLOT-FILLING} and \texttt{INTENT} task and results in the task. CT, RZT, and Coach are from \citet{liu2020coach} and they use 20-shots.} 
       \label{figAPP:NLU}
    \end{minipage}
\end{figure*}

\begin{figure*}[t]
    \begin{minipage}[t]{\textwidth}
      \vspace{0pt}\raggedright
       \centering
        \resizebox{!}{0.05\linewidth}{
        \begin{tabular}{l}
        \rowcolor{mycolor} 
        \begin{tabular}[c]{@{}l@{}}\texttt{i need a cab by 12:30 too $\rightarrow$ leave\_at = 12:30} \\\texttt{i would like the taxi to pick me up from the hotel $\rightarrow$ leave\_at = None}\\\texttt{i would like a taxi from saint john s college $\rightarrow$ leave\_at =}\end{tabular} \\
        \end{tabular}
        }
    \end{minipage}
    \hfill
    \begin{minipage}[t]{\textwidth}
          \vspace{3pt}
        \includegraphics[width=\linewidth]{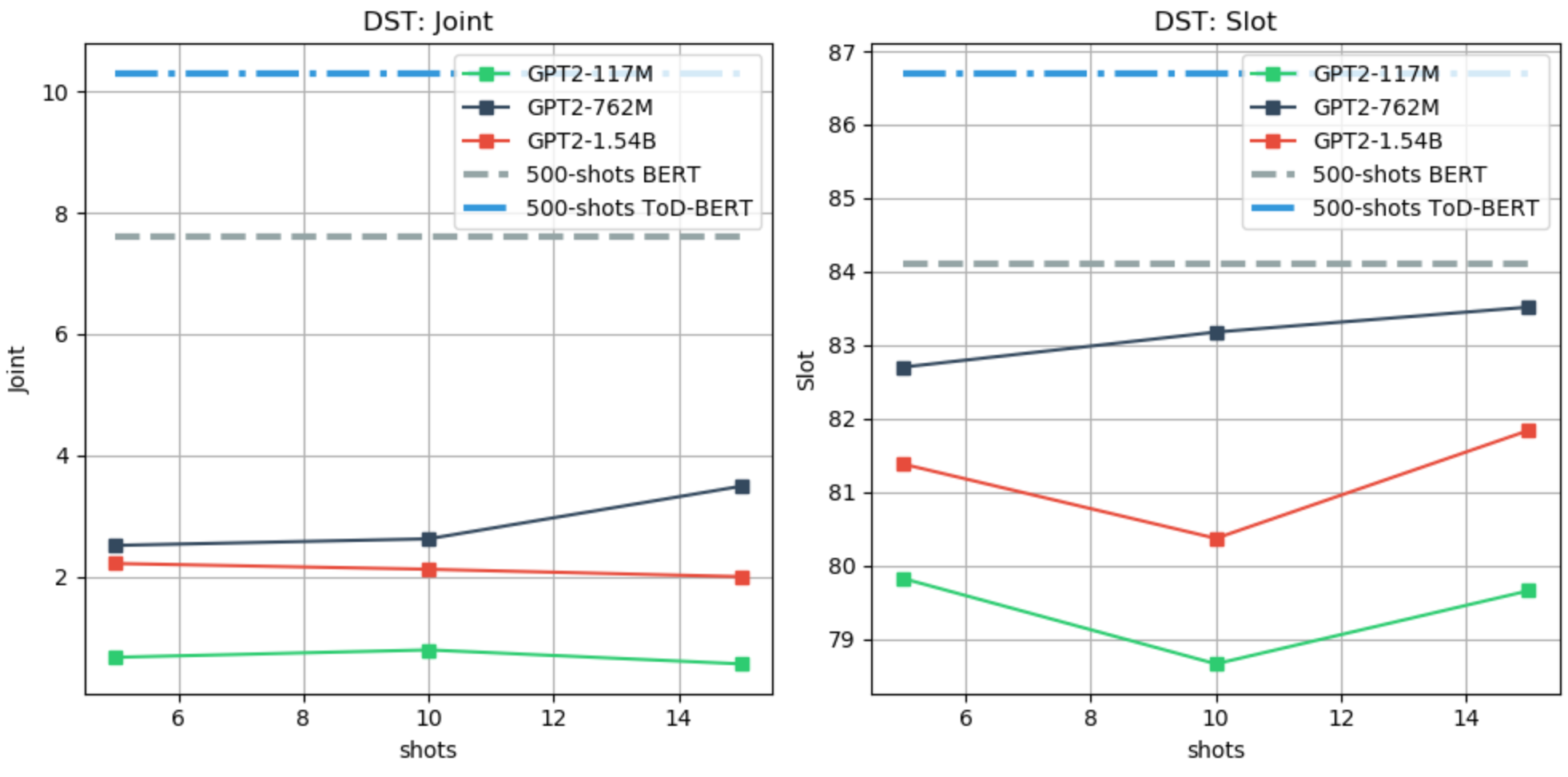}
        \caption{Example of 1-shot LM priming for the \texttt{DST} task and results in the task.  BERT and ToD-BERT are from \citet{wu2020tod} and they use 500 shots.} 
       \label{figAPP:DST}
    \end{minipage}
\end{figure*}

\begin{figure*}[t]
    \begin{minipage}[t]{\textwidth}
      \vspace{0pt}\raggedright
       \centering
        \resizebox{!}{0.04\linewidth}{
        \begin{tabular}{l}
        \rowcolor{mycolor} 
        \begin{tabular}[c]{@{}l@{}}\texttt{yes your booking is successful and your reference number is ri4vvzyc . $\rightarrow$ offerbooked=True} \\\texttt{what type of food are you looking for ? $\rightarrow$ offerbooked=False}\\\texttt{i do not seem to be finding anything $\rightarrow$ offerbooked=}\end{tabular} \\
        \end{tabular}
        }
    \end{minipage}
    \hfill
    \begin{minipage}[t]{\textwidth}
          \vspace{3pt}
          \centering
        \includegraphics[width=0.5\linewidth]{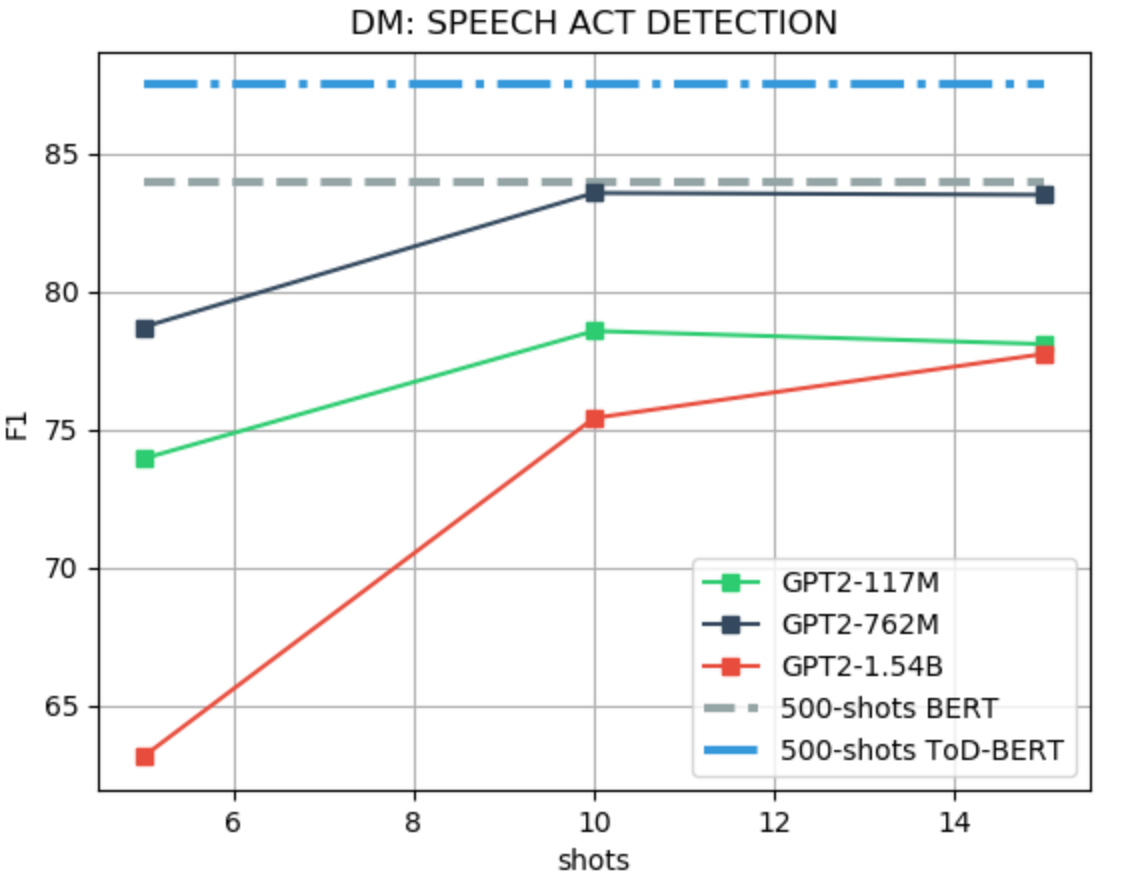}
        \caption{Example of 1-shot LM priming for the \texttt{ACT} task and results in the task.  BERT and ToD-BERT are from \citet{wu2020tod} and they use 500 shots.} 
       \label{figAPP:ACT}
    \end{minipage}
\end{figure*}

\begin{figure*}[t]
    \begin{minipage}[t]{\textwidth}
      \vspace{0pt}\raggedright
       \centering
        \resizebox{!}{0.04\linewidth}{
        \begin{tabular}{l}
        \rowcolor{mycolor} 
        \begin{tabular}[c]{@{}l@{}}\texttt{inform(name=hilton;area=chinatown)$\rightarrow$the hilton is near chinatown} \\\texttt{inform(name=ocean park;phone=4155667020)$\rightarrow$the phone number for ocean park is 4155667020.}\\\texttt{inform(name=super 8 san francisco;phone=8005369326)$\rightarrow$}\end{tabular} \\
        \end{tabular}
        }
    \end{minipage}
    \hfill
    \begin{minipage}[t]{\textwidth}
          \vspace{3pt}
          \centering
        \includegraphics[width=0.9\linewidth]{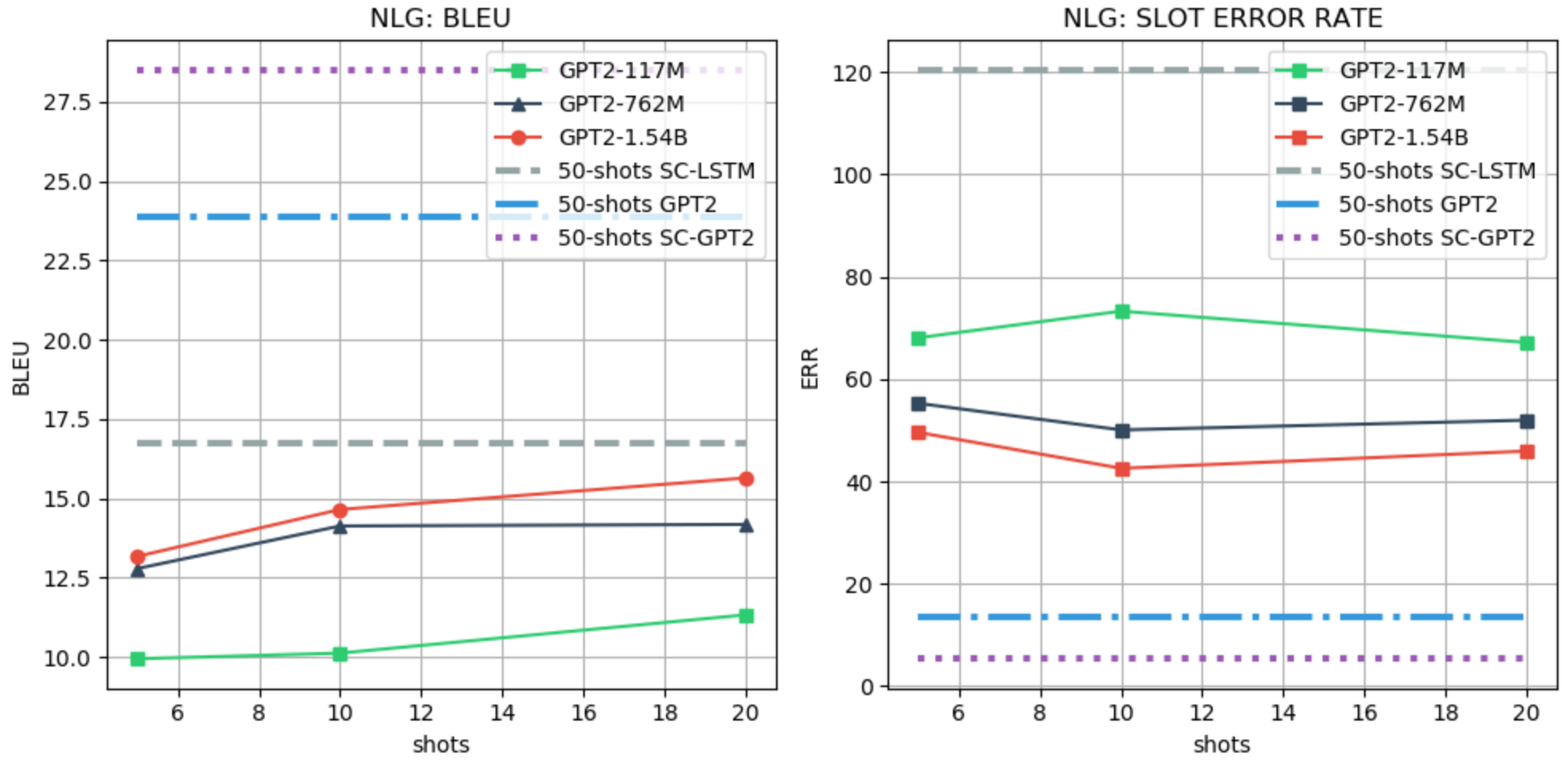}
        \caption{Example of 1-shot LM priming for the \texttt{NLG} task and results in the task. SC-LSTM, GPT-2, and SC-GPT-2 are from \citet{peng2020few}. BLEU the higher the better; SLOT ERROR RATE the lower the better.} 
       \label{figAPP:NLG}
    \end{minipage}
\end{figure*} 	

\begin{table*}[t]
    \centering
    \begin{tabular}{cccccccccc}
    \hline
     Model      &   Shots &   PlayL &   Rest. &   Weather &   PlayM. &   RateBook &   SearchC. &   Find. &     Avg \\
    \hline
     gpt2       &       1 &   31.9008 &    8.0350 &   16.4160 &   32.8150 &   43.6023 &   20.4974 &   24.4994 & 25.3951 \\
     gpt2       &      10 &   46.2546 &   21.6707 &   19.1909 &   21.4724 &   56.2280 &   38.0345 &   41.7234 & 34.9392 \\
     gpt2       &      15 &   54.7410 &   26.4663 &   17.7377 &   28.3369 &   63.8482 &   41.3968 &   47.0525 & 39.9399 \\
     gpt2-large &       1 &   54.7548 &   39.4418 &   23.5223 &   20.8827 &   38.3591 &   26.6576 &   43.0562 & 35.2392 \\
     gpt2-large &      10 &   71.6635 &   39.2936 &   27.7395 &   48.1905 &   61.4562 &   44.4720 &   53.8340 & 49.5213 \\
     gpt2-large &      15 &   71.6569 &   45.5142 &   30.7992 &   46.3439 &   61.7858 &   42.8394 &   61.1420 & 51.4402 \\
     gpt2-xl    &       1 &   53.8250 &   26.2185 &   23.1651 &   28.7647 &   37.1651 &   37.4536 &   31.0224 & 33.9449 \\
     gpt2-xl    &      10 &   70.4698 &   40.5039 &   34.7138 &   40.4731 &   74.3899 &   52.0532 &   64.4166 & 53.8600 \\
     gpt2-xl    &      15 &   67.9448 &   46.9853 &   30.8481 &   44.4646 &   77.1531 &   51.8732 &   67.0917 & 55.1944 \\
    \hline
    \end{tabular}
    \caption{Results in terms of CoNNL F1-score the \texttt{SLOT-FILLING} task.}
    \label{tab:slot-filling}
    \centering
    \begin{tabular}{ccccc}
    \hline
     Model      &   Shots &   Micro &   Macro &     Acc \\
    \hline
     gpt2       &       1 &  0.1600 &  0.1553 & 16.0000 \\
     gpt2       &       2 &  0.1671 &  0.1034 & 16.7143 \\
     gpt2       &       5 &  0.3443 &  0.3223 & 34.4286 \\
     gpt2       &      10 &  0.3600 &  0.3715 & 36.0000 \\
     gpt2-large &       1 &  0.1343 &  0.1188 & 13.4286 \\
     gpt2-large &       2 &  0.3786 &  0.3946 & 37.8571 \\
     gpt2-large &       5 &  0.3300 &  0.3175 & 33.0000 \\
     gpt2-large &      10 &  0.5514 &  0.5871 & 55.1429 \\
     gpt2-xl    &       1 &  0.1700 &  0.1346 & 17.0000 \\
     gpt2-xl    &       2 &  0.3586 &  0.3166 & 35.8571 \\
     gpt2-xl    &       5 &  0.4671 &  0.4371 & 46.7143 \\
     gpt2-xl    &      10 &  0.7300 &  0.7450 & 73.0000 \\
    \hline
    \end{tabular}
    \caption{Results in terms of F1-score (Micro and Macro) and Accuracy in the \texttt{INTENT} recognition task.}
    \label{tab:intent}
\end{table*}

\begin{table*}[t]
    \centering
    \begin{tabular}{ccccc}
    \hline
     Model      &   Shots &   Micro &   Macro &    Acc \\
    \hline
     gpt2       &      5 & 73.9364 & 54.7965 & 0.7394 \\
     gpt2       &      10 & 78.5699 & 59.6442 & 0.7857 \\
     gpt2       &      15 & 78.0943 & 59.8866 & 0.7809 \\
     gpt2-large &      5 & 78.7105 & 62.2181 & 0.7871 \\
     gpt2-large &      10 & 83.5762 & 68.6824 & 0.8358 \\
     gpt2-large &      15 & 83.5102 & 68.2287 & 0.8351 \\
     gpt2-xl    &      5 & 63.1241 & 52.8427 & 0.6312 \\
     gpt2-xl    &      10 & 75.4120 & 62.2672 & 0.7541 \\
     gpt2-xl    &      15 & 77.7434 & 63.0193 & 0.7774 \\
    \hline
    \end{tabular}
    \caption{Results in terms of F1-score (Micro and Macro) and Accuracy in the \texttt{ACT} detection task.}
    \label{tab:act}
\end{table*}

\begin{table*}[t]
    \centering
    \begin{tabular}{cccc}
    \hline
     Model      &   Shots &   Joint &   Slot \\
    \hline
     gpt2       &      5 &     0.7 &   79.8 \\
     gpt2       &      10 &     0.8 &   78.7 \\
     gpt2       &      15 &     0.6 &   79.7 \\
     gpt2-large &      5 &     2.5 &   82.7 \\
     gpt2-large &      10 &     2.6 &   83.2 \\
     gpt2-large &      15 &     3.5 &   83.5 \\
     gpt2-xl    &      5 &     2.2 &   81.4 \\
     gpt2-xl    &      10 &     2.1 &   80.4 \\
     gpt2-xl    &      15 &     2.0 &   81.8 \\
    \hline
    \end{tabular}
    \caption{Results in terms of Joint and Slot Accuracy in the \texttt{DST} task.}
    \label{tab:DST}
\end{table*}

\begin{table*}[]
    \centering
    \begin{tabular}{cccccccccc}
                    
    \hline
     Model         & Shots &   restaurant &   laptop &   hotel &    tv &   attraction &   train &   taxi &   Avg \\
    \hline
     SC-LSTM       &  50    & 15.90 &    21.98 &   31.30 & 22.39 &         7.76 &    6.08 &  11.61 &  16.71 \\
     GPT-2         &  50    & 29.48 &    27.43 &   35.75 & 28.47 &        16.11 &   13.72 &  16.27 &  23.89 \\
     SC-GPT        &  50    & 38.08 &    32.73 &   38.25 & 32.95 &        20.69 &   17.21 &  19.70 &  28.51 \\ \hline
     gpt2        & 5&        9.93 &    17.75 &   14.85 & 16.29 &         5.50 &    0.26 &   5.01 &  9.94 \\
     gpt2       &    10&       8.10 &    17.75 &   16.85 & 16.29 &         5.84 &    1.30 &   4.71 & 10.12 \\
     gpt2       &  20&      10.68 &    17.75 &   19.15 & 16.29 &         4.89 &    3.24 &   7.28 & 11.32 \\
     gpt2-large  & 5&        10.60 &    24.42 &   13.92 & 24.58 &         7.38 &    0.73 &   7.86 & 12.78 \\
     gpt2-large &       10&    13.10 &    24.42 &   20.68 & 24.58 &         6.68 &    3.18 &   6.25 & 14.13 \\
     gpt2-large &   20&      11.47 &    24.42 &   16.13 & 24.58 &         7.97 &    5.30 &   9.36 & 14.18 \\
     gpt2-xl     & 5&        13.65 &    23.39 &   14.26 & 26.61 &         6.96 &    0.74 &   6.59 & 13.17 \\
     gpt2-xl    &    10&       14.51 &    23.39 &   19.42 & 26.61 &         8.21 &    4.00 &   6.40 & 14.65 \\
     gpt2-xl    &    20&     17.02 &    23.39 &   21.30 & 26.61 &         6.43 &    5.68 &   9.06 & 15.64 \\
    \hline
    \end{tabular}
    \caption{Results in terms of BLEU score for the \texttt{NLG} task. SC-LSTM, GPT-2, and SC-GPT-2 are from \citet{peng2020few}.}
    \label{tab:NLG_BLEU}
    
    \begin{tabular}{cccccccccc}
    \hline
     Model          & Shots &   restaurant &   laptop &   hotel &    tv &   attraction &   train &   taxi &   Avg \\
    \hline
     SC-LSTM        & 50    & 48.02 &    80.48 &   31.54 & 64.62 &       367.12 &  189.88 &  61.45 &  120.44 \\
     GPT-2          & 50    & 13.47 &    11.26 &   11.54 &  9.44 &        21.10 &   19.26 &   9.52 &  13.65 \\
     SC-GPT         & 50    &  3.89 &     3.39 &    2.75 &  3.38 &        12.72 &    7.74 &   3.57 &  5.35 \\\hline
     gpt2        & 5&       60.48 &    60.84 &   73.63 & 72.66 &        81.79 &   60.54 &  66.67 & 68.09 \\
     gpt2       & 10&      72.75 &    60.84 &   78.02 & 72.66 &        80.49 &   88.75 &  59.52 & 73.29 \\
     gpt2       & 20&       70.36 &    60.84 &   74.18 & 72.66 &        67.20 &   68.96 &  55.95 & 67.16 \\
     gpt2-large  & 5&       55.39 &    36.33 &   84.62 & 44.02 &        64.31 &   58.11 &  44.05 & 55.26 \\
     gpt2-large & 10&         57.49 &    36.33 &   62.09 & 44.02 &        52.31 &   73.27 &  25.00 & 50.07 \\
     gpt2-large & 20&       48.20 &    36.33 &   85.71 & 44.02 &        56.07 &   61.35 &  32.14 & 51.98 \\
     gpt2-xl     & 5&       44.61 &    29.99 &   67.03 & 37.92 &        67.63 &   55.82 &  44.05 & 49.58 \\
     gpt2-xl    & 10&         46.41 &    29.99 &   47.80 & 37.92 &        50.87 &   62.36 &  22.62 & 42.57 \\
     gpt2-xl    & 20& 44.61 &    29.99 &   68.68 & 37.92 &        56.50 &   52.93 &  30.95 & 45.94 \\
    \hline
    \end{tabular}
    \caption{Results in terms of SLOT ERROR RATE for the \texttt{NLG} task. SC-LSTM, GPT-2, and SC-GPT-2 are from \citet{peng2020few}.}
    \label{tab:NLG_SLR}
\end{table*}


\end{document}